\documentclass[11pt,a4paper]{article}
\usepackage{acl2019,times}

\usepackage{url}
\usepackage{booktabs}       % professional-quality tables
\usepackage{amsfonts}       % blackboard math symbols
\usepackage{nicefrac}       % compact symbols for 1/2, etc.
\usepackage{microtype}      % microtypography
\usepackage{amsmath}
\usepackage{bbm}
\usepackage{color}
\usepackage{multirow}
\usepackage{graphicx}
\usepackage{caption}
\usepackage{subcaption}
\usepackage{amssymb}
\usepackage{enumitem}
\usepackage{makecell}
\usepackage[symbol]{footmisc}
\usepackage{float}
\usepackage{mwe,tikz}
\usepackage[percent]{overpic}

\newcommand{\expect}{\mathbb{E}}
\renewcommand{\eqref}[1]{Eq.~(\ref{#1})}

\renewcommand{\cite}{\citep}

\usepackage{pifont}% http://ctan.org/pkg/pifont
\newcommand{\cmark}{\ding{51}}%
\newcommand\indicator[1]{\mathbbm{1}[#1]}

\aclfinalcopy

\title{Stay on the Path: Instruction Fidelity in Vision-and-Language Navigation}
\author{Vihan Jain\thanks{\ Authors contributed equally.} \quad 
        Gabriel Magalhaes\footnotemark[\value{footnote}] \quad 
        Alexander Ku\footnotemark[\value{footnote}] \quad
        Ashish Vaswani \\
        \textbf{Eugene Ie} \quad
        \textbf{Jason Baldridge} \\
  Google Research \\
  {\tt \{vihan, gamaga, alexku, avaswani, eugeneie, jridge\}@google.com}
  }
  
\newcommand{\dataext}{R4R}
\newcommand{\smallpad}{\vspace{.2cm}}

\begin{document}

\maketitle

\begin{abstract}
Advances in learning and representations have reinvigorated work that connects language to other modalities. A particularly exciting direction is Vision-and-Language Navigation (VLN), in which agents interpret natural language instructions and visual scenes to move through environments and reach goals. Despite recent progress, current research leaves unclear how much of a role language understanding plays in this task, especially because dominant evaluation metrics have focused on goal completion rather than the sequence of actions corresponding to the instructions. Here, we highlight shortcomings of current metrics for the Room-to-Room dataset \cite{Anderson:2018:VLN} and propose a new metric, Coverage weighted by Length Score (CLS). We also show that the existing paths in the dataset are not ideal for evaluating instruction following because they are direct-to-goal shortest paths. We join existing short paths to form more challenging extended paths to create a new data set, Room-for-Room (R4R). Using R4R and CLS, we show that agents that receive rewards for instruction fidelity outperform agents that focus on goal completion.

\end{abstract}

% ---------------------------------------------------------------------
% Math Notation
% ---------------------------------------------------------------------
% $a$ (scalar)
% $\textbf{a}$ (vector)
% $A = (a_1, \cdots, a_n)$ (tuple)
% $\mathcal{A} = \{a_1, \cdots, a_n\}$ (set)
% $\textbf{A}$ (matrix)
% $\alpha_i = \text{Softmax}(\textbf{a})_i$
% $\textbf{x} = \text{Attention}(\textbf{q}, \textbf{K}, \textbf{V})$
% $\textbf{X} = \text{Attention}(\textbf{Q}, \textbf{K}, \textbf{V})$
% $\textbf{a} = \text{Attention}(\textbf{q}, \textbf{K})$
% $\textbf{A} = \text{Attention}(\textbf{q}, \textbf{K})$
% $Y = \text{LSTM}(X)$
% ---------------------------------------------------------------------

\section{Introduction}
\label{sec:intro}

In Vision-and-Language Navigation (VLN) tasks, agents must follow natural language navigation instructions through either simulated \cite{Macmahon06walkthe,Yan:18,bisk:etal:2018,follownet}, simulations of realistic \cite{Blukis:18visit-predict,misra-etal-2018-mapping} and real environments \cite{Anderson:2018:VLN,DBLP:journals/corr/abs-1807-03367,Chen19:touchdown,cirik2018following}, or actual physical environments \cite{skovaj:etal:2016,thomason:aaai18,WilliamsGRT18}. Compared to other tasks involving co-grounding in visual and language modalities -- such as image and video captioning \cite{7558228,7298754,Vinyals2015ShowAT,Wang2018VideoCV,YuCVPR2016}, visual question answering (VQA) \cite{7410636,Yang2016StackedAN}, and visual dialog \cite{visdial} -- VLN additionally requires agents to plan their actions, move, and dynamically respond to changes in their visual field. 

%The concordance of neural representations and learning for both language and vision has greatly facilitated research into problems that integrate these modalities. A variety of neural architectures can be applied to both vision and language, and they operate on the same core representations and can be straightforwardly integrated and jointly trained in the same model when multimodal data is available. The language and vision components can be independently trained on problems from each modality and then tuned to the multimodal data, thereby facilitating transfer and reducing the data requirements for new multimodal tasks. As such, there has been a flowering in the last few years of new multimodal problems and datasets (cite,cite,cite). 

\begin{figure}
  \centering   
  %\begin{overpic}[clip, trim=2cm 9cm 2cm 8cm, width=\linewidth]{imgs/stay_on_path_noisy.pdf}
  %   \put(22,62){\includegraphics[width=0.5\linewidth]{imgs/cover_metrics.png}}  
  %\end{overpic}
 \includegraphics[clip, trim=5cm 15cm 5cm 11cm, width=\linewidth]{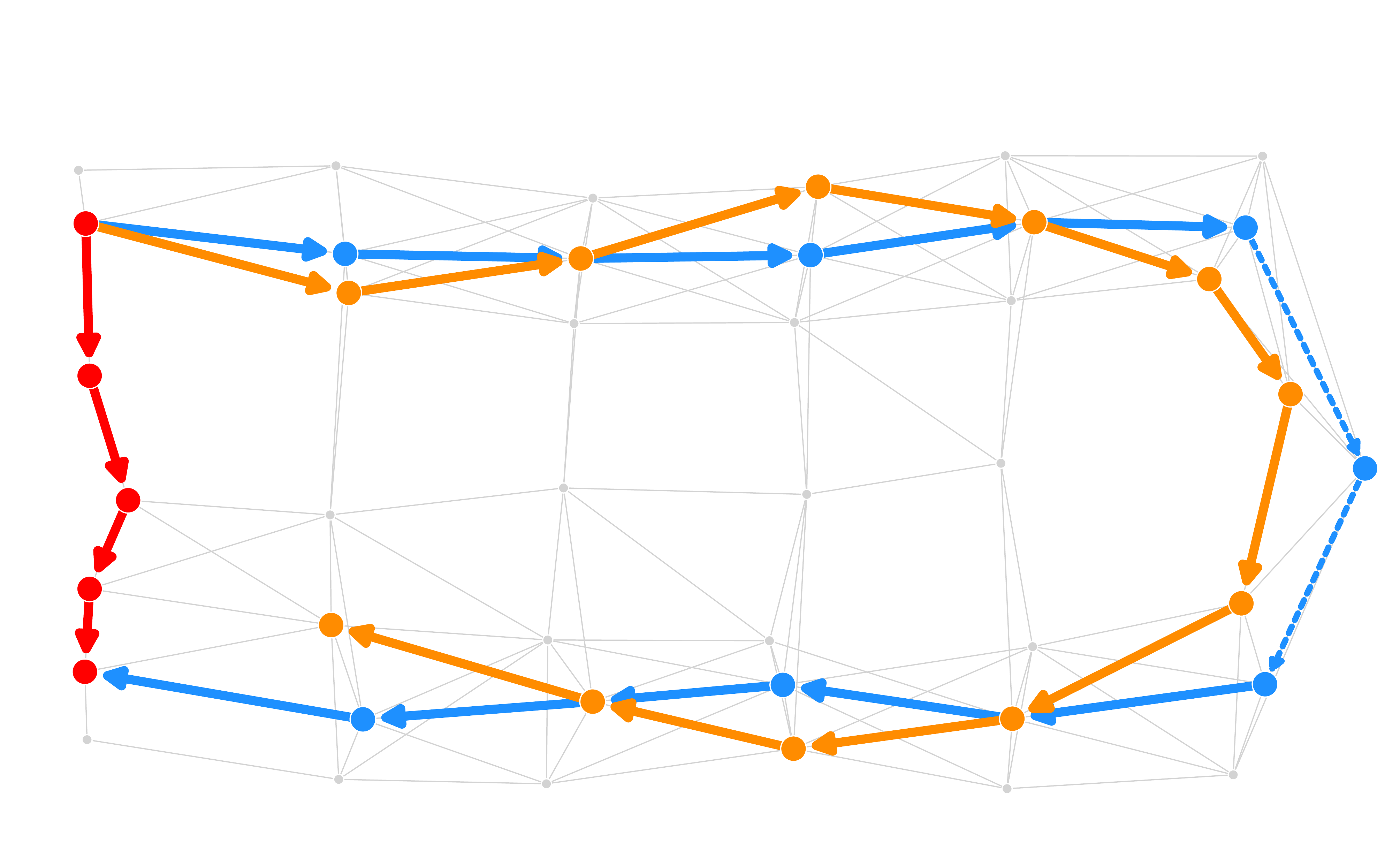}
%      \includegraphic
\caption{
    \textit{It's the \textbf{journey}, not just the goal.}
    To give language its due place in VLN, we compose paths in the R2R dataset to create longer, twistier R4R paths (blue). Under standard metrics, agents that head straight to the goal (red) are not penalized for ignoring the language instructions: for instance, SPL yields a perfect 1.0 score for the red and only 0.17 for the orange path. In contrast, our proposed CLS metric measures fidelity to the reference path, strongly preferring the agent with the orange path (0.87) over the red one (0.23).  
%     % To give language its due place in R2R, we compose paths to create longer, twistier R4R paths (blue). Agents that head straight to the goal (red) get a perfect SPL score, but one which follows the path (orange) gets low SPL. Unlike dominant metrics, CLS strongly prefers the orange agent's path.
    }
    \label{fig:spl_sucks}
\end{figure}

% TODO: Cite the Points, Paths and Playscapes paper here in final version, if accepted. 

Photo-realistic simulations for VLN are especially promising: they retain messy, real world complexity and can draw on pre-trained models and rich data about the world, but do not require investment in and maintenance of physical robots and spaces for them. Given this, we focus on the Room-to-Room (R2R) task \cite{Anderson:2018:VLN}. Despite significant recent progress on R2R since its introduction \cite{Fried:2018:Speaker,Ma:2019:SelfMonitoringAgent,Wang:2018:RCM}, the structure of the dataset and current evaluation metrics greatly diminish the importance of language understanding for the task. The core problems are that paths in R2R are all direct-to-goal shortest paths and metrics are mostly based on goal completion rather than fidelity to the described path. To address this, we define a new metric, Coverage weighted by Length Score (CLS), and compose path pairs of R2R to create Room-for-Room (R4R), an algorithmically produced extension of R2R. Figure \ref{fig:spl_sucks} illustrates path composition and the scores of two agent paths for both CLS and Success weighted by Path Length (SPL), a metric recently proposed by \citet{Anderson:2018:Evaluation}. In the example, an agent which ignores the language but gets to the goal receives a perfect SPL score.

Language is not irrelevant for R2R. \citet{thomason:naacl19} ablate visual and language inputs and find that withholding either from an action sampling agent reduces performance on unseen houses. Also, the generated instructions in the augmented paths of \citet{Fried:2018:Speaker} improved performance for several models. However, while many of these augmented instructions have clear starting \textit{or} ending descriptions, the middle portions are often disconnected from the path they are paired with (see \citet{huang-etal-splu-robonlp-2019} for in depth analysis of augmented path instructions). That these low-fidelity augmented instructions improve results indicates that current metrics are insensitive to instruction fidelity. Our new CLS metric measures how closely an agent's trajectory conforms with the entire reference path, not just goal completion.

Because the reference paths in R2R are all direct-to-goal, the importance of the actual journey taken from start to finish is diminished; as a result, fidelity between instructions and their corresponding paths is harder to evaluate. In longer, twistier paths, the importance of not always going directly to the goal becomes much clearer. We take advantage of the fact that the original R2R data contains many paths that have goals that coincide with the start points of other paths. By concatenating pairs of paths and their corresponding instructions, we create longer paths that allow us to better gauge the ability of an agent to stick to the path as described. With this data, Reinforced Cross-modal Matching models \cite{Wang:2018:RCM} that use CLS as a reward signal dramatically improve not only CLS (from 20.4\% for the agent  with goal-oriented rewards to 34.6\%), but navigation error also reduces from 8.45m to 8.08m on the the Validation Unseen dataset. Furthermore, we find that the agent with goal-oriented rewards obtains the same CLS (20.4) on \dataext\ regardless of whether the full instruction or only the last five tokens are provided to it. In contrast, the CLS-rewarded agent drops from CLS of 34.6 to 25.3 when given only the last five tokens.

%We present three main contributions. First, we show R2R is not ideal for evaluating the role of language and extend it to a richer sets of paths better suited for measuring instruction fidelity. Second, we point out drawbacks of standard VLN metrics and propose a novel metric, CLS, capable of measuring how a predicted trajectory conforms to the reference one. Finally, we show that CLS can be included in the reward system of an RL agent, outperforming agents trained with goal-oriented metrics and showing greater fidelity to the path described by the instructions.

\section{Extending R2R to create R4R}
\label{sec:extended}

Instructions such as ``\textit{Turn left, walk up the stairs. Enter the bathroom.}" are easy for people but challenging for computational agents. Agents must segment instructions, set sub-goals based on understanding them and ground the language and their actions in real world objects and dynamics. An agent may need expectations for how spatial scenes change when turning. Additionally, it must recognize visual and environmental features that indicate it has entered or encountered something referred to as ``\textit{the bathroom}" and know to stop.

\subsection{Room-to-Room (R2R)}

Room-to-Room (R2R) supports visually-grounded natural language navigation in a photo-realistic environment \cite{Anderson:2018:VLN}. R2R consists of an environment and language instructions paired to reference paths. The environment defines a graph where nodes are possible positions an agent may inhabit. Edges indicate that a direct path between two nodes is navigable. For each node, R2R provides an egocentric panoramic view. All images are collected from buildings and house interiors. The paths paired with language instructions are composed by sequences of nodes in this graph.

For data collection, starting and goal nodes are sampled from the graph and the shortest path between those nodes is taken, provided it is no shorter than 5m and contains between 4 and 6 edges. Each path has 3 associated natural language instructions, with an average length of 29 words and a total vocabulary of 3.1k words. Apart from the training set, the dataset includes two validation sets and a test set. One of the validation sets includes new instructions on environments overlapping with the training set (Validation Seen), and the other is entirely disjoint from the training set (Validation Unseen).

\begin{figure}
\centering
\includegraphics[angle=90, clip, trim=6cm 0cm 6cm 0cm, width=0.8\linewidth]{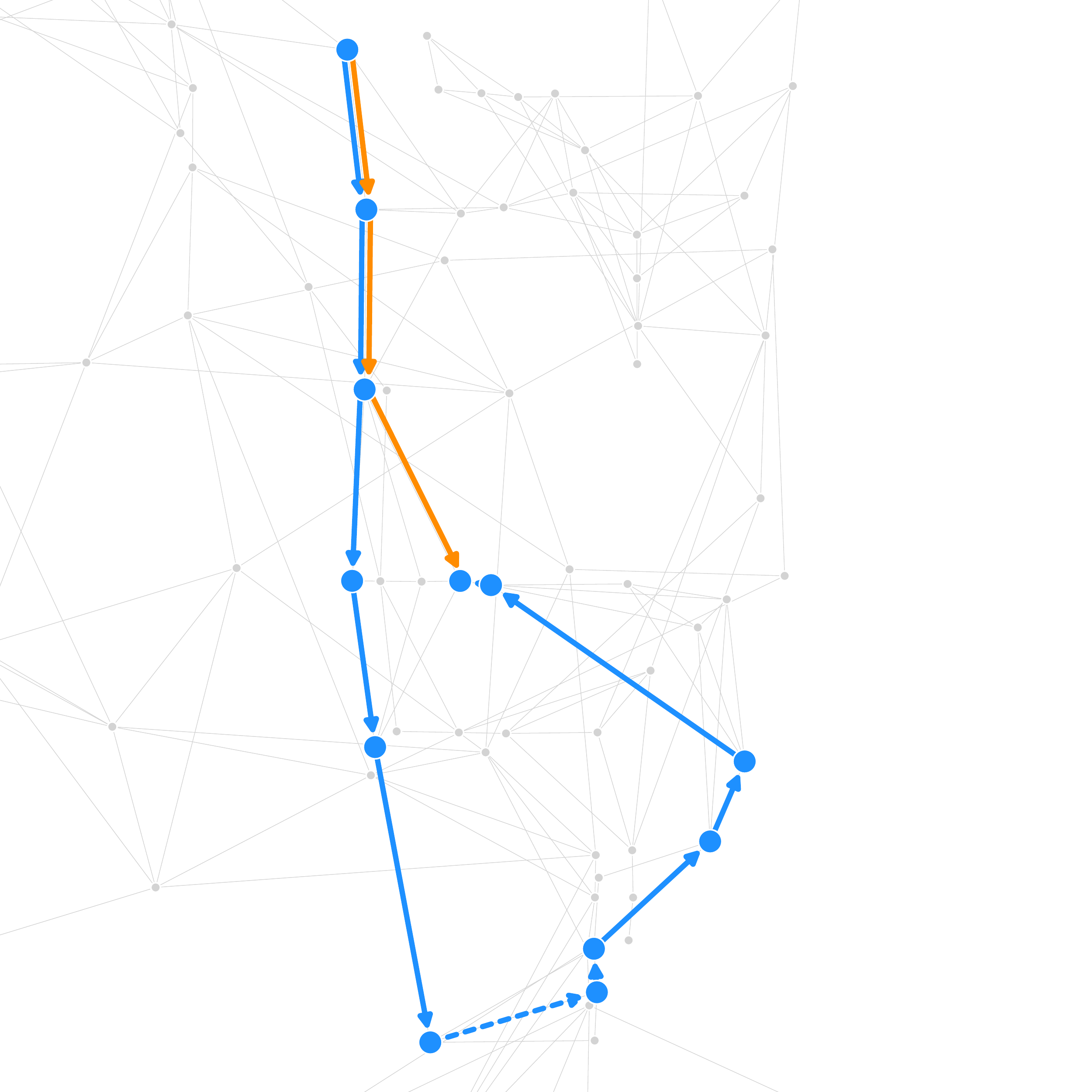}
\caption{An example of an extended path in the \dataext{} dataset, where the dotted blue arrow connects two blue paths with solid arrows, corresponding to the instructions \textit{``Make a left down at the narrow hall beside the office and walk straight to the exit door. Go out the door and wait."} and \textit{``Turn around and enter the bedroom. Walk to the other side of the room and turn left. Walk into the doorway leading out and stop."}. The shortest-to-goal path from the starting point is shown in orange. \label{fig:r4r-example}}
\end{figure}

\citet{Fried:2018:Speaker} propose a \textit{follower} model which is trained using student forcing, where actions are sampled from the agent's decisions, but supervised using the action that takes the agent closest to the goal. During inference, the \textit{follower} generates candidate paths which are then scored by a \textit{speaker} model. The \textit{speaker} model was also used for creating an augmented dataset that is used as an extension of training data by the \textit{follower} model as well as by many subsequently published models. \citet{Wang:2018:RCM}  train their agents using policy gradients. At every step, the agent is rewarded for getting closer to the target location (extrinsic reward) as well as for choosing an action that reduces cycle-reconstruction error between instruction generated by a matching critic and ground-truth instruction (intrinsic reward). In both papers, there is little analysis presented about the generative models.

%There are outstanding concerns about how well agents trained on one task can transfer their learnings to another task, \eg, how quickly (if at all) can an agent trained on the indoor R2R dataset can adapt to a different dataset, such as Touchdown \cite{Chen19:touchdown}, which is set in an outdoor environment. More 

Recently, \citet{Anderson:2018:Evaluation} pointed out weaknesses in the commonly used metrics for evaluating the effectiveness of agents trained on these tasks. A new metric, Success weighted by Path Length (SPL) was proposed that penalized agents for taking long paths. Any agent using beam search (e.g. \citet{Fried:2018:Speaker}), is penalized heavily by this metric. There have also been concerns about structural biases present in these datasets which may provide \textit{hidden} shortcuts to agents training on these problems. \citet{thomason:naacl19} presented an analysis on R2R dataset, where the trained agent continued to perform surprisingly well in the absence of language inputs. 
%We also question the biases of these datasets and the effectiveness of current metrics.

% - Importance of augmented data \citet{Fried:2018:Speaker,Ma:2019:SelfMonitoringAgent} in getting good results. From hereon out we denote R2R as both the 

\subsection{Room-for-Room (\dataext)}
\label{subsec:r2r2}\label{subsec:r4r}

Due to the process by which the data are generated, all R2R reference paths are shortest-to-goal paths. 
%As such, they capture only a small subset of the richness of navigation. 
Because of this property, conformity to the instructions is \textbf{decoupled} from reaching the desired destination -- and this short-changes the language perspective. In a broader scope of reference paths, the importance of following language instructions in their entirety becomes clearer, and proper evaluation of this conformity can be better studied. Additionally, the fact that the largest path in the dataset has only 6 edges exacerbates the challenge of properly evaluating conformity. This motivates the need for a dataset with larger and more diverse reference paths.

% This prohibits us from properly evaluating the properties of CLS, since paths that yield a high CLS will naturally yield a high SPL metric as well. R2R does not contain reference paths where SPL is high and CLS is low, which occurs when the reference path has winding turns and an agent learns to take the shortest path; nor does it contain paths where SPL is low ans CLS is high, which occurs when the again fails to reach the goal but otherwise closely follows the reference path.

\begin{table}
\setlength\tabcolsep{4.8pt}
\begin{tabular}{llccc}
 &  & \#samples  & $\text{PL}(R)$ & $d(r_1, r_{|R|})$ \\\Xhline{2\arrayrulewidth}
\multirow{3}{*}{R2R} & Train       & 14039 & 9.91 & 9.91  \\
                     & Val. seen   & 1021  & 10.2 & 10.2 \\
                     & Val. unseen & 2249  & 9.50 & 9.50 \\\hline
\multirow{3}{*}{\dataext} & Train       & 233613 & 20.6 & 10.5  \\
                          & Val. seen   & 1035   & 20.4 & 11.1 \\
                          & Val. unseen & 45162  & 20.2 & 10.1
%                          \\\Xhline{2\arrayrulewidth}
% N (R2R)           & 14039  & 1021      & 2349        \\
% N (R4R)  & 233613 & 1035      & 45162       \\
% PL, DL        & 9.91   & 10.19     & 9.50        \\
% PL (extended) & 20.60  & 20.37     & 20.21       \\
% DL (extended) & 10.50  & 11.11     & 10.05       \\
\end{tabular}
\caption{Comparison of R2R to \dataext{}. $\text{PL}(R)$ represents the mean path length of the reference paths and $d(r_1, r_{|R|})$ is mean length of the shortest-to-goal path.\label{tab:dataset-compare}}
\end{table}

\begin{figure*}
\centering
\includegraphics[width=.99\linewidth]{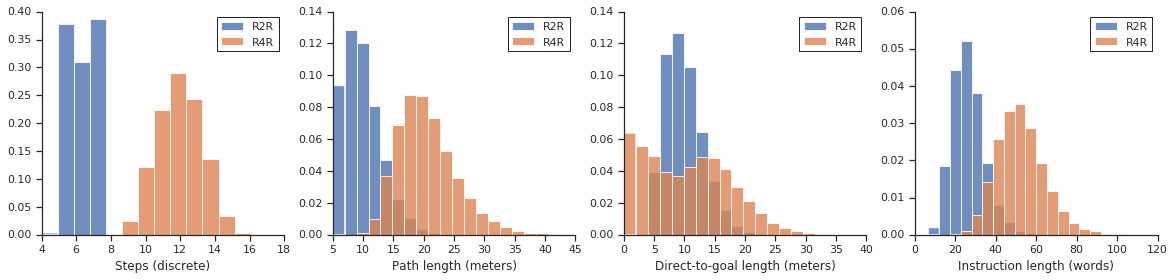}
\caption{From left to right, the distribution of the number of steps, path lengths, direct-to-goal path lengths and instruction lengths in the original R2R and extended R4R datasets.\label{fig:data-distributions}}
% \caption{Distribution of path lengths (left) and instruction lengths (right) in the original and extended datasets.\label{fig:data-distributions}}
\end{figure*}

To address the lack of path variety, we propose a data augmentation strategy\footnote{R2R-to-R4R code is at {\footnotesize \href{https://github.com/google-research/google-research/tree/master/r4r}{https://github.com/google-research/google-research/tree/master/r4r}}} that introduces long, twisty paths without additional human or low-fidelity machine annotations (e.g. those from \citet{Fried:2018:Speaker}). Existing paths in the dataset can be extended by joining them with other paths that start within some threshold of where they end. Formally, two paths $A{=}(a_1, a_2, \cdots, a_{|A|})$ and $B{=}(b_1, b_2, \cdots, b_{|B|})$ are joined if $d(a_{|A|}, b_1){<}d_{th}$. The resulting extended paths are thus $R{=}(a_1, \cdots, a_{|A|}, c_1, \cdots, c_{|C|}, b_1, \cdots, b_{|B|})$, where $C = (c_1, c_2, \cdots, c_{|C|})$ is the shortest path between $a_{|A|}$ and $b_1$. (If $a_{|A|}{=}b_1$, $C$ is empty.)

Each combination of instructions corresponding to paths $A$ and $B$ is included in \dataext. Since each path maps to multiple human-annotated instructions, each extended path will map to $N_A \cdot N_B$ joined instructions, where $N_A$ and $N_B$ are the number of annotations associated with paths $A$ and $B$, respectively. Figure \ref{fig:r4r-example} shows an example of an extended path and the corresponding instructions, compared to the shortest-to-goal path.

% \alexku{TODO: Discuss how this is a cheap way to increase compositionality on the language side. ALso talk about the distribution of instruction lengths -- though this isn't that interesting...}

%\alexku{Define what R2R (raw data + speaker augmented) and R4R (just the concatenated) is. }

% \begin{table}
% \begin{tabular}{l|rrr}
%               & Train  & Val. Seen & Val. Unseen \\
% \hline
% N (R2R)           & 14039  & 1021      & 2349        \\
% N (R4R)  & 233613 & 1035      & 45162       \\
% PL, DL        & 9.91   & 10.19     & 9.50        \\
% PL (extended) & 20.60  & 20.37     & 20.21       \\
% DL (extended) & 10.50  & 11.11     & 10.05       \\
% \end{tabular}
% \caption{Table comparing the extended dataset ($d_{th} = 3.0$) to the original dataset, where PL and DL are the mean path lengths and Dijkstra lengths, respectively, and N are the number of path-instruction pairs per training split.\label{tab:dataset-compare}}
% \end{table}

\section{Evaluation Metrics in VLN}
\label{sec:metrics}

Historically, the performance of VLN models has been evaluated with respect to the objective of reaching the goal location. The nature of the path an agent takes, however, is of clear practical importance: it is undesirable for any robotic agent in the physical world to reach the destination by taking a different path than what it was instructed to follow; failure to comply with instructions might lead to navigating unwanted and potentially dangerous locations. Here, we propose a series of desiderata for VLN metrics and introduce Coverage weighted by Length Score (CLS).  Table \ref{tab:nav-metrics} provides a high level summary of this section's contents.

\newcommand{\lastP}{\ensuremath{p_{|P|}}}
\newcommand{\lastR}{\ensuremath{r_{|R|}}}

\subsection{Desiderata}
\label{subsec:desiderata}

%One critical aspect of navigation tasks is performance evaluation, where a comparison of predicted and reference paths is needed. 
Commonly, navigation tasks are defined in a discrete space: the environment determines a graph where each node is a position the agent could be in and each edge between two nodes represents that there is a navigable step between them. Let the predicted path $P = (p_1, p_2, p_3, ..., \lastP)$ be the sequence of nodes visited by the agent and reference path $R = (r_1, r_2, r_3, ..., \lastR)$ be the sequence of nodes in the reference trajectory. Generally, $p_1 = r_1$, since in many VLN tasks, the agent begins at the reference path's start node. The following desiderata characterize metrics that gauge the fidelity of $P$ with respect to $R$ rather than just goal completion. Throughout the paper, we refer to the subsequent desired properties as \textit{Desideratum} (i). 

% \begin{itemize}
%     \setlength{\itemsep}{0em}
%     \item[\textbf{P0}] Metrics should be designed to measure a notion of similarity between a predicted path $P$ and a reference path $R$, which should depend on all nodes in $P$ and all nodes in  $R$.
%     \item[\textbf{P1}] Metrics should penalize differences from the reference path according to a soft notion of discrepancy.
%     \item[\textbf{P2}] Metrics should yield a perfect score if and only if the two paths are an exact match.
%     \item[\textbf{P3}] Metrics should be invariant to scale.
%     \item[\textbf{P4}] Metrics should be computationally tractable.
% \end{itemize}

\bgroup
\def\arraystretch{1.3}
\begin{table*}
\centering
\setlength\tabcolsep{3.8pt}
\scalebox{0.9}{
\begin{tabular}{lclccccc}\Xhline{2\arrayrulewidth}

\multirow{2}{4cm}{Metric}     & \multirow{2}{0.5cm}{$\uparrow$ $\downarrow$} & \multirow{2}{2cm}{Definition}                          &  \multicolumn{5}{c}{Desiderata coverage} \\
                              &              &                                                                                        &  (1)   &  (2)   &  (3)   &  (4)   &  (5)    \\\Xhline{2\arrayrulewidth}
Path Length (PL)              &      -       & $\sum_{1 \le i < |P|} d(p_i, p_{i+1})$                                                 &        & \cmark &        &        & \cmark  \\
% Success Rate (SR)             & $\uparrow$   & $\indicator{d(p_{|P|}, r_{|R|}) \le d_{th}}$                                           &        &        &        & \cmark & \cmark  \\
% Oracle Success Rate (OSR)     & $\uparrow$   & $\indicator{ \exists p \in P \mid d(p, r_{|R|}) \le d_{th}}$                           &        &        &        & \cmark & \cmark  \\ 
Navigation Error (NE)         & $\downarrow$ & $d(p_{|P|}, r_{|R|})$                                                                  &        & \cmark &        &        & \cmark  \\
Oracle Navigation Error (ONE) & $\downarrow$ & $\min_{p \in P}d(p, r_{|R|})$                                                          &        & \cmark &        &        & \cmark  \\
Success Rate (SR)             & $\uparrow$   & $\indicator{\text{NE}(P, R) \le d_{th}}$                                               &        &        &        & \cmark & \cmark  \\
Oracle Success Rate (OSR)     & $\uparrow$   & $\indicator{\text{ONE}(P, R) \le d_{th}}$                                              &        &        &        & \cmark & \cmark  \\ 
Success weighted by PL (SPL)  &  $\uparrow$   & $\text{SR}(P,R) \cdot \dfrac{d(p_1, r_{|R|})}{\max\{\text{PL}(P), d(p_1, r_{|R|})\}}$ &        & \cmark &        & \cmark & \cmark  \\
Success weighted by Edit Distance (SED)         & $\uparrow$ & $\text{SR}(P,R) \left(1 - \dfrac{\text{ED}(P, R)}{\max{\{|P|, |R|}\} - 1}\right)$        & \cmark & & \cmark & \cmark & \cmark         \\\hline
Coverage weighted by LS (CLS)      & $\uparrow$   & $\text{PC}(P, R) \cdot \text{LS}(P, R)$                               & \cmark  & \cmark & \cmark & \cmark & \cmark  \\\Xhline{2\arrayrulewidth}
\end{tabular}
}
\caption{Definition and desiderata coverage of navigation metrics. \label{tab:nav-metrics}}
\end{table*}
\egroup

\noindent
\textbf{(1)} \textit{Path similarity measure.} Metrics should characterize a notion of similarity between a predicted path $P$ and a reference path $R$. This implies that metrics should depend on all nodes in $P$ and all nodes in $R$, which contrasts with many common metrics which only consider the last node in the reference path (see Section \ref{subsec:existing_metrics}). Metrics should penalize deviations from the reference path, even if they lead to the same goal. This is not only prudent, as agents might wander around undesired terrain if this is not enforced, but also explicitly gauges the fidelity of the predictions with respect to the provided language instructions.

\noindent
\textbf{(2)} \textit{Soft penalties.} Metrics should penalize differences from the reference path according to a soft notion of dissimilarity that depends on distances in the graph. This ensures that larger discrepancies are penalized more severely than smaller ones and that metrics should not rely only on dichotomous views of intersection. For instance, a predicted path that has no intersection to the reference path, but follows it closely, as illustrated in Figure \ref{fig:spl_sucks} should not be penalized too severely.

\noindent
\textbf{(3)} \textit{Unique optimum.} Metrics should yield a perfect score if and only if the reference and predicted paths are an exact match. This ensures that the perfect score is unambiguous: the reference path $R$ is therefore treated as a golden standard. No other path should have the same or higher score as the reference path itself.

\noindent
\textbf{(4)} \textit{Scale invariance.} Metrics should be consistent over different datasets.

\noindent
\textbf{(5)} \textit{Computational tractability.} Metrics should be pragmatic, allowing fast automated evaluation of performance in navigation tasks.

\subsection{Existing Navigation Metrics}
\label{subsec:existing_metrics}

Table \ref{tab:nav-metrics} defines previous navigation metrics and how they match our desiderata. We denote by $d(n, m)$ the shortest distance between two nodes along the edges of the graph and $d(n, P) = \min_{p \in P} d(n, p)$ the shortest distance between a node and a path. All distances are computed along the edges of the graph determined by the environment, which are not necessarily equal to the euclidean distance between the nodes.

\textit{Path Length (PL)} measures the total length of the predicted path, which has the optimal value equal to the length of the reference path. \textit{Navigation Error (NE)} measures the distance between the last node in the predicted path and the last reference path node. \textit{Oracle Navigation Error (ONE)} measures the shortest distance from any node in the predicted path to the last reference path node. \textit{Success Rate (SR)} measures how often the last node in the predicted path is within a threshold distance $d_{th}$ of the last reference path node. \textit{Oracle Success Rate (OSR)} measures how often any node in the predicted path is within a threshold distance $d_{th}$ of the last node in the reference path.

\textit{Success weighted by Path Length (SPL)} \cite{Anderson:2018:Evaluation} takes into account both Success Rate and the normalized path length. It was proposed as a single summary measure for navigation tasks. Note that the agent should maximize this metric, and it is only greater than 0 if the success criteria was met. While this metric is ideally suited when the evaluating whether the agent successfully reached the desired destination, it does not take into account any notion of similarity between the predicted and reference trajectories and fails to take into account the intermediary nodes in the reference path. As such, it violates \textit{Desideratum} (1). Since there could exist more than one path with optimal length to the desired destination, it also violates \textit{Desideratum} (3).
    
\begin{figure}
\centering
\includegraphics[width=0.9\linewidth]{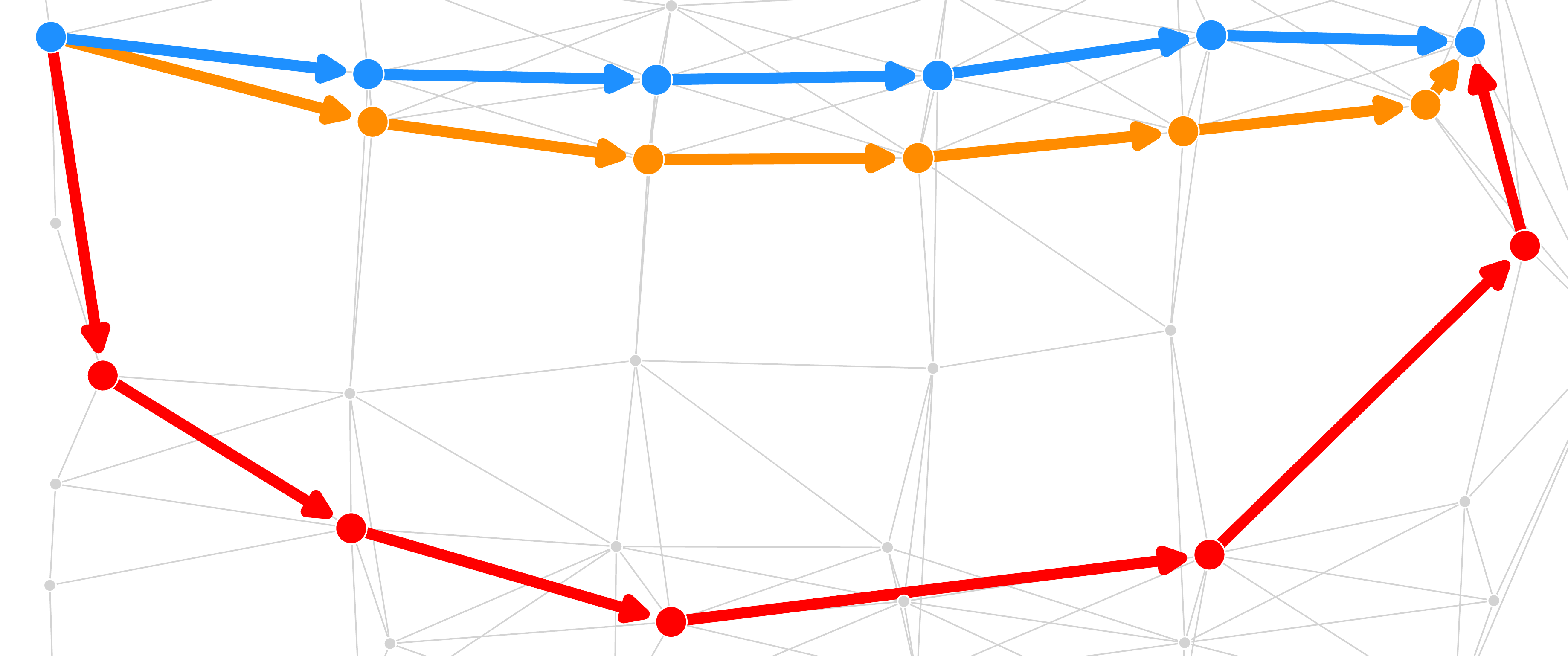}
\caption{With respect to the blue path, SED yields zero for both the orange and red paths, while CLS yields a score of 0.89 for orange and 0.48 for red. \label{fig:desideratum2}}
\end{figure}

\textit{Success weighted by Edit Distance (SED)} \cite{Chen19:touchdown} is based on the edit distance $\text{ED}(P,R)$ between the two paths, equal to the Levenshtein distance between the two sequences of actions $A_P = ((p_1, p_2), (p_2, p_3), ..., (p_{|P|-1}, p_{|P|}))$ and $A_R = ((r_1, r_2), (r_2, r_3), ..., (r_{|R|-1}, r_{|R|}))$. The Levenshtein distance is the minimum number of edit operations (insertion, deletion and substitution of actions) that can transform path $A_R$ into $A_P$. Similarly to SPL, SED is also multiplied by $\text{SR}(P,R)$, so only paths that meet the success criteria receive a score greater than 0. This metric naturally satisfies \textit{Desideratum} (1), (3) and (4). Further, it is possible to compute it using dynamic programming in $O(|P||R|)$, further satisfying \textit{Desideratum} (5). \textit{Desideratum} (2), however, is left unsatisfied, as SED does not take into account \textit{how} two actions differ from each other (considering, for instance, the graph distance between their end nodes), but only if they are the same or not. This subtle but important difference is illustrated in Figure \ref{fig:desideratum2}.

%\footnotetext[1]{See Equations \ref{eq:cls}- \ref{eq:ls}.}

\subsection{Coverage weighted by Length Score}
\label{subsec:cls}

% \gamaga{We should consider open sourcing code for evaluating this metric.}

We introduce \textit{Coverage weighted by Length Score (CLS)} as a single summary measure for VLN. CLS is the product of the Path Coverage (PC) and Length Score (LS) of the agent's path $P$ with respect to reference path $R$: 

\begin{equation}
    \text{CLS}(P, R) = \text{PC}(P, R) \cdot \text{LS}(P, R)
    \label{eq:cls}
\end{equation}

PC replaces SR as a non-binary measure of how well the reference path is covered by the agent's path.  It is the average coverage of each node in the reference path $R$ with respect to path $P$:

\begin{equation}
    \text{PC}(P, R) = \dfrac{1}{|R|} \sum_{r \in R}\exp\left(-\dfrac{d(r, P)}{d_{th}}\right)
    \label{eq:pc}
\end{equation}

\noindent
where $d(r, P){=}\min_{p \in P} d(r, p)$ is the distance to reference path node $r$ from the nearest node in $P$. The coverage contribution for each node $r$ is an exponential decay of this distance. ($1/d_{th}$ is a decay constant to account for graph scale.)

LS compares the predicted path length $\text{PL}(P)$ to EPL, the expected optimal length given $R$'s coverage of $P$. If say, the predicted path covers only half of the reference path (i.e., $PC = 0.5$), then we expect the optimal length of the predicted path to be half of the length of the reference path. As a result, EPL is given by:

\begin{equation}
\text{EPL}(P, R) = \text{PC}(P, R) \cdot \text{PL}(R)
\end{equation}

LS for a predicted path $P$ is optimal only if $\text{PL}(P)$ is equal to the expected optimal length -- it is penalized when the predicted path length is shorter or longer than the expected path length:

\begin{equation}
\text{LS}(P, R) = \dfrac{\text{EPL}(P, R)}{\text{EPL}(P, R) + |\text{EPL}(P, R) - \text{PL}(P)|}
                % &= \dfrac{\text{PC} \cdot \text{PL}(R)}{\text{PC} \cdot \text{PL}(R) + |\text{PC} \cdot \text{PL}(R) - \text{PL}(P)|}
                \label{eq:ls}
\end{equation}

There is a clear parallel between the terms of CLS and SPL. CLS replaces success rate, the first term of SPL, with path coverage, a continuous indicator for measuring how well the predicted path covered the nodes on the reference path. Unlike SR, PC is sensitive to the intermediary nodes in the reference path $R$. The second term of SPL penalizes the path length $\text{PL}(P)$ of the predicted path against the optimal (shortest) path length $ d(p_1, r_{|R|})$; CLS replaces that with length score LS, which penalizes the agent path length $\text{PL}(P)$ against EPL, the expected optimal length for its coverage of $R$.

% Figure \ref{fig:metrics_comp} shows a comparison between statistics on the two metrics on the R2R and \dataext{} datasets (see Section \ref{sec:extended}), using 100,000 samples with randomly chosen reference paths from the datasets. Predicted paths are obtained by a random policy with uniform probability of choosing between the neighbors of a node and to stop. On all sampled predicted paths, the agent starts at the beginning of the reference path.

% \begin{figure*}
%     \centering
%      \includegraphics[width=0.8\linewidth]{imgs/cls_spl_dist.pdf}
%     \qquad
%     \scriptsize
%     \caption{Comparisons between CLS and SPL metrics on random paths inside houses on R2R and \dataext{} datasets (see Section \ref{sec:extended}).}
%     \label{fig:metrics_comp}
% \end{figure*}

CLS naturally covers \textit{Desideratum} (1) and (2). Assuming that the reference path is acyclic and that $p_1 = r_1$, i.e., reference and predicted path start at the same node, \textit{Desideratum} (3) is also satisfied. Additionally, CLS also covers \textit{Desideratum} (4) because PC and LS are both invariant to the graph scale (due to the term $d_{th}$). Finally, the distances from each pair of nodes in the graph can be pre-computed using Dijkstra's algorithm \cite{dijkstra1959note} for each node, resulting in a complexity of $O(EV + V^2\log(V))$, where $V$ and $E$ are the number of vertices and edges in the graph, respectively. $PC(P, R)$ can be computed in $O(|P||R|)$, and $LS(P, R)$ can be computed in $O(|P| + |R|)$, making CLS satisfy \textit{Desideratum} (5).

% As a disclaimer, this metric is intended to be used only on environments where the sampling granularity on which we discretize the space does not vary much within the graph (i.e., there is not much variance in the lengths of the edges). If there are regions of the environment where sampling is denser, those points will undesirably weight more on computing the average coverage $\text{PC}(P,R)$.

%Finally, we show sample comparisons of SPL and CLS on random paths inside houses on R2R dataset in Figure \ref{fig:spl_vs_cls}.

% \begin{figure*}
%      \centering
%       \includegraphics[clip, trim=0cm 9cm 0cm 8cm, width=\linewidth]{imgs/spl_vs_cls.pdf}
%      \qquad
%      \scriptsize
%      \caption{Comparisons between CLS and SPL metrics on random paths inside houses on R2R dataset.}

%      \label{fig:spl_vs_cls}
%\end{figure*}

\section{Agent}
\label{sec:model}

We reimplement the Reinforced Cross-Modal Matching (RCM) agent of \citet{Wang:2018:RCM} and extend it to use a reward function based on both CLS (Section \ref{subsec:cls}) as well as success rate.

\subsection{Navigator}
\label{sec:navigator}

The \textit{reasoning navigator} of \citet{Wang:2018:RCM} learns a policy $\pi_{\theta}$ over parameters $\theta$ that map the natural language instruction $\mathcal{X}$ and the initial visual scene $v_1$ to a sequence of actions $a_{1..T}$.
At time step $t$, the agent state is modeled using a LSTM \cite{Hochreiter:1997:LSTM} that encodes the trajectory of past visual scenes and agent actions, $h_t{=}LSTM([v_t; a_{t-1}], h_{t-1})$, where $v_t$ is the output of visual encoder as described below. 

\smallpad
\noindent
\textbf{Language Encoder} Language instructions $\mathcal{X}=x_{1..n}$ are initialized with pre-trained GloVe word embeddings \cite{Pennington:2014:GloVe} that are fine-tuned during training. We restrict the GloVe vocabulary to tokens that occur at least five times in the instruction data set. All out of vocabulary tokens are mapped to a single OOV identifier. Using a bidirectional recurrent network \cite{Schuster:1997:BiRNN} we encode the instruction into language contextual representations $w_{1..n}$.

\smallpad
\noindent
\textbf{Visual Features}
As in \citet{Fried:2018:Speaker}, at each time step $t$, the agent perceives a 360-degree panoramic view of its surroundings from the current location. The view is discretized into $m$ view angles ($m=36$ in our implementation, 3 elevations x 12 headings at 30-degree intervals).
The image at view angle $i$, heading angle $\phi$ and elevation angle $\theta$ is represented by a concatenation of the pre-trained CNN image features with the 4-dimensional orientation feature [sin $\phi$; cos $\phi$; sin $\theta$; cos $\theta$] to form $v_{t,i}$.
The visual encoder pools the representation of all view angles $v_{t,1..m}$ using attention over the previous agent state $h_{t-1}$.
\begin{equation}
v_t = \text{Attention}(h_{t-1}, v_{t,1..m}) 
\label{eq:v_t}
\end{equation}

\noindent
The actions available to the agent at time $t$ are denoted as $u_{t,1..l}$, where $u_{t,j}$ is the representation of navigable direction $j$ from the current location obtained similarly to $v_{t,i}$ \cite{Fried:2018:Speaker}. The number of available actions, $l$, varies for different locations, since nodes in the graph have different number of connections.

\smallpad
\noindent
\textbf{Action Predictor}
As in \citet{Wang:2018:RCM}, the model predicts the probability $p_k$ of each navigable direction $k$ using a bilinear dot product.

\begin{small}
\begin{align}
p_k &= \text{softmax}([h_t; c_t^{text}; c_t^{\text{visual}}]W_c(u_{t,k}W_u)^T)\\
c_t^{\text{text}} &= \text{Attention}(h_t, w_{1..n}) \\
c_t^{\text{visual}} &= \text{Attention}(c_t^{\text{text}}, v_{t, 1..m})
\end{align}
\end{small}

\subsection{Learning}
\label{sec:learning}

Training is performed using two separate phases, (1) behavioral cloning \cite{Bain:1999:Cloning,Wang:2018:RCM,Daftry:2016:TransferablePolicies} and (2) REINFORCE policy gradient updates \cite{Williams:1992:PolicyGradient}.

As is common in cases where expert demonstrations are available, the agent's policy is initialized using behavior cloning to constrain the learning algorithm to first model state-action spaces that are most relevant to the task, effectively warm starting the agent with a good initial policy. No reward shaping is required during this phase as behavior cloning corresponds to solving the following maximum-likelihood problem,

\begin{equation}
\max_{\theta} \sum_{(s,a) \in \mathcal{D}} \log \pi_{\theta}(a|s)
\end{equation}

\noindent
where $\mathcal{D}$ is the demonstration data set.

After warm starting the model with behavioral cloning, we obtain standard policy gradient updates by sampling action sequences from the agent's behavior policy. As in standard policy gradient updates, the model is optimized by minimizing the loss function $\mathcal{L}^{\text{PG}}$ whose gradient is the negative policy gradient estimator \cite{Williams:1992:PolicyGradient}.

\begin{equation}
\mathcal{L}^{\text{PG}} = -\hat{\expect}_t[\log \pi_{\theta}(a_{t}|s_{t}) \hat{A}_t]
\end{equation}

\noindent
where the expectation $\hat{\expect}_t$ is taken over a finite batch of sample trajectories generated by the agent's stochastic policy $\pi_{\theta}$. To reduce variance, we scale the gradient using the advantage function $\hat{A}_t{=} R_t{-}\hat{b}_t$. ($R_t{=}\sum_{i=t}^\infty \gamma^{i-t}r_i$ is the observed $\gamma$-discounted episodic return and $\hat{b}_t$ is the estimated value of the agent's current state at time $t$.)

The models are trained using mini-batch gradient descent. Our experiments show that interleaving behavioral cloning and policy gradient training phases improves performance on the validation set. Specifically we interleaved each policy gradient update batch with $K$ behaviour cloning batches, with the value of $K$ decaying exponentially, such that the training strategy asymptotically becomes only policy gradient updates.

\subsection{Reward}
\label{sec:reward}

For consistency with the established benchmark \cite{Wang:2018:RCM}, we implemented a dense \textit{goal-oriented} reward function that optimizes the \textit{success rate} metric. This includes an immediate reward at time step $t$ in an episode of length $T$, given by:

\begin{equation}
\label{eq:denseSRreward}
r(s_t, a_t) =
    \begin{cases}
        d(s_t, r_{|R|}) - d(s_{t+1}, r_{|R|}) & \text{if $t < T$} \\
        \indicator{d(s_T, r_{|R|}) \leq d_{th}} & \text{if $t = T$}
    \end{cases}
\end{equation}

\noindent
where $d(s_t, r_{|R|})$ is the distance between $s_t$ and target location $r_{|R|}$, $\indicator{\cdot}$ is the indicator function, $d_{th}$ is the maximum distance from $r_{|R|}$ that the agent is allowed to terminate for success.

To incentivize the agent to not only reach the target location but also to conform to the reference path, we also train our agents with following \textit{fidelity-oriented} sparse reward:

\begin{equation}
\label{eq:sparseSRCLSreward}
r(s_t, a_t) =
    \begin{cases}
        0 & \text{if $t < T$} \\
        \indicator{d(s_T, r_{|R|}) \leq d_{th}} + \\
        CLS(s_{1...T}, R) & \text{if $t = T$}
    \end{cases}
\end{equation}

\noindent
where $R$ is the reference path in the dataset associated with the instruction $\mathcal{X}$. This rewards actions that are consistent both with reaching the goal and following the path corresponding to the language instructions. It is worth noting here that, similar to Equation \ref{eq:denseSRreward}, a relative improvement in CLS can be added as a reward-shaping term for time steps $t < T$, however empirically we did not find noticeable difference in the performance of agents trained with or without the shaping term. For simplicity, all of the experiments involving \textit{fidelity-oriented} reward use the sparse reward in Equation \ref{eq:sparseSRCLSreward}.

\section{Results}
\label{sec:results}

We obtain the performance of models trained under two training objectives. The first is \textit{goal oriented} (Equation \ref{eq:denseSRreward}): agents trained using this reward are encouraged to pursue only the last node in the reference path. The second is \textit{fidelity oriented} (Equation \ref{eq:sparseSRCLSreward}): agents trained using this reward receive credit not only for reaching the target location successfully but also for conforming to the reference path. We report the performance on standard metrics (PL, NE, SR, SPL) as well as the new CLS metric.

To further explore the role of language, we perform ablation studies, where agents are trained using the full language instructions and evaluated on partial (last 5 tokens) or no instructions. With no instructions, the agent only has the full visual input, similar to the unimodal ablation studies of \citet{thomason:naacl19}. To eliminate the effect observed due to distribution shift during evaluation and preserve the length distribution of the input instructions, we further conducted studies where agents are given arbitrary instructions from the validation set, with the reference path remaining unaltered. We observed that experiments with arbitrary instruction had similar results to studies where instructions where fully removed.

\footnotetext[2]{Our \textit{goal oriented} results match the RCM benchmark on validation unseen but are lower on validation seen. We suspect this is due to differences in implementation details and hyper-parameter choices.}
\footnotetext[3]{For the random evaluation, we first sample the number of edges in the trajectory from the distribution of number of edges in the reference paths of the training dataset. Then, for each node, we uniformly sample between its neighbors and move the agent there. We report the average metrics for 1 million random trajectories.}
\footnotetext[4]{As in \citet{Wang:2018:RCM}, we report the performance of Speaker-Follower model from \citet{Fried:2018:Speaker} that utilizes panoramic action space and augmented data but no beam search (pragmatic inference) for a fair comparison.}
\footnotetext[5]{We report the performance of the RCM model without intrinsic reward as the benchmark.}

\begin{table*}[h]
\centering
\def\arraystretch{1.04}
\setlength\tabcolsep{3.3pt}
\scalebox{0.9}{
\begin{tabular}{cp{5.2cm}ccccc@{\hskip 0.4cm}ccccc}\Xhline{2\arrayrulewidth} \\[-0.8em]
                              &  & \multicolumn{5}{c}{\textbf{Validation Seen}} & \multicolumn{5}{c}{\textbf{Validation Unseen}}\\\cmidrule(lr{0.2cm}){3-7}\cmidrule{8-12}
\# & Model                            & PL    & NE $\downarrow$   & SR $\uparrow$   & SPL $\uparrow$   & \underline{CLS} $\uparrow$  & PL    & NE $\downarrow$   & SR $\uparrow$   & SPL $\uparrow$   & \underline{CLS} $\uparrow$\\\Xhline{2\arrayrulewidth}
0 & Random\footnote[3]                   & 10.4  & 9.82              & 5.0             & 3.7              & 29.4            &  9.32  & 9.32              & 5.2             & 4.0              & 29.0            \\
1 & Speaker-Follower \cite{Fried:2018:Speaker}\footnote[4]         & -     & 3.36              & 66.4            & -                & -               &  -     & 6.62              & 35.5            & -                & -            \\
2 & RCM \cite{Wang:2018:RCM}\footnote[5] & 12.1  & \textbf{3.25}     & \textbf{67.6}   & -                & -               &  15.0  & 6.01              & 40.6            & -                & -               \\\hline

3 & Speaker-Follower              & 15.5  & 4.98              & 50.1            & 40.1             & 54.8            &  15.2  & 6.36              & 35.3            & 28.1             & 42.9            \\
4 & RCM, \textit{goal oriented }         & 13.7  & 4.48              & 55.3            & 47.9             & \textbf{61.1}   &  14.8  & \textbf{6.00}     & \textbf{41.1}   & 32.7             & 47.4 \\
5 & $~~~$ last 5 tokens                  & 16.9  & 7.35              & 26.5            & 22.2             & 39.0            &  15.1  & 8.16              & 22.2            & 17.2             & 35.1             \\
6 & $~~~$ no instructions                & 21.1  & 7.78              & 22.3            & 11.6             & 27.5            &  17.7  & 8.69              & 13.0            & 9.4              & 26.1             \\   
7 & RCM, \textit{fidelity oriented}      & 12.2  & 4.63              & 57.3            & \textbf{50.7}    & 60.2            &  13.2  & 6.38              & 40.8            & \textbf{35.1}    & \textbf{50.9}   \\
8 & $~~~$ last 5 tokens                  & 13.4  & 8.08              & 27.8            & 23.5             & 42.4            &  14.4  & 8.29              & 23.2            & 17.7             & 35.5             \\
9 & $~~~$ no instructions                & 20.1  & 8.95              & 18.2            & 8.8              & 24.8            &  20.5  & 8.76              & 14.3            & 6.2              & 22.7             \\ \Xhline{2\arrayrulewidth}
\end{tabular}
}
\caption{Results on R2R Validation Seen and Validation Unseen sets. Rows 0 and 3-9 shows numbers from our implementations. SR, SPL and CLS are reported as percentages and NE and PL in meters. \label{tab:r2r-original-paths-results}}
\end{table*}

\begin{table*}[h]
\centering
\def\arraystretch{1.04}
\setlength\tabcolsep{3.3pt}
\scalebox{0.9}{
\begin{tabular}{cp{5.2cm}ccccc@{\hskip 0.4cm}ccccc}\Xhline{2\arrayrulewidth} \\[-0.8em]
                              &  & \multicolumn{5}{c}{\textbf{Validation Seen}} & \multicolumn{5}{c}{\textbf{Validation Unseen}}\\\cmidrule(lr{0.2cm}){3-7}\cmidrule{8-12}
\# & Model                            & PL    & NE $\downarrow$   & SR $\uparrow$   & SPL $\uparrow$   & \underline{CLS} $\uparrow$  & PL    & NE $\downarrow$   & SR $\uparrow$   & SPL $\uparrow$   & \underline{CLS} $\uparrow$\\\Xhline{2\arrayrulewidth}     
0 & Random\footnote[3]               & 21.8  & 11.4              & 13.1            & 2.0              & 23.1            & 23.6  & 10.4              & 13.8            & 2.2              & 22.3  \\
1 & Speaker-Follower                 & 15.4  & 5.35              & 51.9            & \textbf{37.3}    & 46.4            & 19.9  & 8.47              & 23.8            & \textbf{12.2}    & 29.6 \\
2 & RCM, \textit{goal oriented}      & 24.5  & \textbf{5.11}     & \textbf{55.5}   & 32.3             & 40.4            & 32.5  & 8.45              & \textbf{28.6}   & 10.2             & 20.4\\
3 & $~~~$ last 5 tokens              & 29.5  & 8.73              & 26.4            & 12.4             & 35.1            & 29.5  & 9.04              & 23.4            & 4.5              & 20.4\\
4 & $~~~$ no instructions            & 32.3  & 9.50              & 20.7            & 8.0              & 33.3            & 34.0  & 9.45              & 19.0            & 2.3              & 17.4 \\   
5 & RCM, \textit{fidelity oriented}  & 18.8  & 5.37              & 52.6            & 30.6             & \textbf{55.3}   & 28.5  & \textbf{8.08}     & 26.1            & 7.7              & \textbf{34.6} \\
6 & $~~~$ last 5 tokens              & 17.1  & 8.88              & 24.8            & 11.7             & 39.3            & 25.5  & 8.52              & 18.9            & 5.6              & 25.3 \\
7 & $~~~$ no instructions            & 12.7  & 10.5              & 12.1            & 5.4              & 37.2            & 22.8  & 9.41              & 15.5            & 4.9              & 23.0 \\   \Xhline{2\arrayrulewidth}
\end{tabular}
}
\caption{Results on R4R Validation Seen and Validation Unseen sets (see Section \ref{sec:extended}). SR, SPL and CLS are reported as percentages and NE and PL in meters. \label{tab:r2r-extended-paths-results}}
\end{table*}

On the \dataext\ dataset, the \textit{fidelity oriented} agent significantly outperforms the \textit{goal oriented} agent ($>14$\% absolute improvement in CLS), demonstrating that including CLS in the reward signal successfully produces better conformity to the reference trajectories. Furthermore, on Validation Unseen, when all but the last 5 tokens of instructions are removed, the \textit{goal oriented} agent yields the same CLS as with the full instructions, while the \textit{fidelity oriented} agent suffers significantly, decaying from $34.6\%$ to $25.3\%$. This indicates that including fidelity measurements as reward signals improve the agent's reliance on language instructions--thereby better keeping the \textbf{L} in VLN.

%%%%%% @Jason: Please review this as you may have additional comments!!

\subsection{R2R Performance}

Table \ref{tab:r2r-original-paths-results} summarizes the experiments on R2R.\footnotemark[2] There are not major differences between \textit{goal oriented} and \textit{fidelity oriented} agents, highlighting the problematic nature of R2R paths with respect to instruction following: essentially, rewards that only take into account the goal implicitly signals path conformity---by the construction of the dataset itself. As a result, an agent optimized to reach the target destination may incidentally appear to be conforming to the instructions. The results shown in Section \ref{subsec:r4r_results} further confirm this hypothesis by training and evaluate \textit{goal oriented} and \textit{target oriented} agents on \dataext{} dataset.

As evidenced by the ablation studies, models draw some signal from the language instructions. However, having the last five tokens makes up for a significant portion of the gap between no instructions and full instructions, again highlighting problems with R2R and the importance in R2R of identifying the right place to stop rather than following the path. The performance of both the agents degrade in similar proportions when instructions are partially or fully removed. 
%We note here that on R2R, it is difficult to distinguish an agent that is more carefully attending to the instructions from an agent that is not. This is due to an inherent bias in the R2R dataset that agents are able to exploit: all the paths in the dataset are shortest-to-goal paths.

Finally, as expected, the SPL metric appears consistent with CLS on R2R, since all reference paths are shortest-to-goal. As highlighted in Section \ref{subsec:r4r_results}, this breaks in settings where paths twist and turn.

% \begin{itemize}
%     \item Our reimplementations of Speaker-Follower and RCM are worse on seen, but match them on unseen -- hypothesis is that our visual features have less capacity, so the agents in our implementation are less able to memorize.
%     \item The gap between SF and RCM for CLS is interesting (42.9 to 50.9, also holding for val seen). Is there a story here?

%     \item On R2R, the SPL metric appears consistent with CLS. If you only have shortest-paths, it's fine. (But trouble awaits when we get twisty paths!)
%     \item There are not major differences between goal-oriented and fidelity-oriented agents on R2R, high-lighting the problematic nature of the paths in the dataset with respect to instruction following: essentially, optimizing SR on R2R is implicitly optimizing CLS by construction of the dataset.
%     \item Instructions do matter! However, having the last five tokens makes up for a lot of the gap between no instructions and full instructions, again high-lighting problems with R2R and the ability to identify a good place to stop rather than following the path.
%     \item On R2R, the SPL metric appears consistent with CLS. If you only have shortest-paths, it's fine. (But trouble awaits when we get twisty paths!)
% \end{itemize}

\subsection{R4R Performance}
\label{subsec:r4r_results}
% Key observations:

% \begin{itemize}
%     \item The fidelity reward shows a massive improvement for CLS, for both seen and unseen!
%     \item The goal oriented agent still gets better NE on val seen, but the tables turn for val unseen. Even though goal agent has a higher SR on unseen, it has a higher variance. It is wandering farther (higher PL) and still getting to a pretty good stopping point, but it goes clearly goes astray.
%     \item SF actually beats RCM wrt CLS! Presumably this is because it is supervised per step rather than rewarded for goal completion?
%     \item SPL goes crazy on R4R: utterly useless. It ranks goal oriented agent higher (as expected) than the fidelity agent.
%     \item Language ablations show the expected behavior: massive reduction in CLS for fidelity agent, but almost no change for goal agent.
% \end{itemize}

Table \ref{tab:r2r-extended-paths-results} shows the results on \dataext{}. Overall, the scores for all model variants on R4R are much lower than R2R, which highlights the additional challenge of following longer instructions for longer paths. Most importantly, the \textit{fidelity oriented} agent significantly outperforms the \textit{goal oriented} agent for both CLS and navigation error, demonstrating the importance of both measuring path fidelity and using it to guide agent learning.

On the experiments, the \textit{goal oriented} agent continues to exploit biases and the underlying structure in the environment to reach the goal. When the instructions are removed during evaluation, the agent's performance on the CLS metric barely degrades, showing that the agent does not rely significantly on the instructions for its performance. In contrast, the \textit{fidelity oriented} agent learns to pursue conformity to the reference path, which in turn requires attending more carefully to the instructions. When instructions are removed during evaluation, performance of the \textit{fidelity oriented} agent degrades considerably on the CLS metric. In fact, the \textit{fidelity oriented} agent performs better on CLS metric without instructions as the \textit{goal oriented} agent performs with the full instructions.

Furthermore, we highlight that historically dominant metrics are ineffective -- even misleading -- for measuring agents' performance: for instance, especially for reference paths that begin and end at close locations, SPL is a poor measure of success since it assumes the optimal path length is the shortest distance between the starting and ending positions (as illustrated in Figure \ref{fig:spl_sucks}, for example). This is particularly noticeable from the results: the \textit{goal oriented} agent gets better SPL scores than the \textit{fidelity oriented} agent, even when it has massively poorer performance on conformity (CLS). 

% \footnotetext[?]{Note that results are reported on models trained using augmented data, as in \cite{Wang:2018:RCM}, for a fair comparison.}
% NOTE: footnote 5,6,7 are defined in conclusion.tex

%\input{related}

\section{Conclusion}
\label{sec:conclusion}

The CLS metric, \dataext, and our experiments provide a better toolkit for measuring the impact of better language understanding in VLN. Furthermore, our findings suggests ways that future datasets and metrics for judging agents should be constructed and set up for evaluation. The R4R data itself clearly still has considerable headroom: our reimplementation of the RCM model gets only 34.6 CLS on paths in R4R's Validation Unseen houses. Keeping in mind that humans have an average navigation error of 1.61 in R2R \cite{Anderson:2018:VLN}, the average navigation error of 8.08 meters for R4R by our best agent leaves plenty of headroom. Future agents will need to make effective use of language and its connection to the environment to both drive CLS up and bring NE down in R4R.

We expect path fidelity to not only be interesting with respect to grounding language, but to be crucial for many VLN-based problems. For example, future extensions of VLN will likely involve games \cite{baldridge-etal-2018-points} where the instructions being given take the agent around a trap or help it avoid opponents. Similar constraints could hold in search-and-rescue human-robot teams \cite{tradr-project,tradr-italy}, where the direct path could take a rolling robot into an area with greater danger of collapse. In such scenarios, going straight to the goal could be literally deadly to the robot or agent.

\section*{Acknowledgments}

We would like to thank our anonymous reviewers and the Google Research team, especially Radu Soricut, for the insightful comments that contributed to this paper.

% \bibliography{mybib}

\begin{thebibliography}{}
\expandafter\ifx\csname natexlab\endcsname\relax\def\natexlab#1{#1}\fi

\bibitem[{Anderson et~al.(2018{\natexlab{a}})Anderson, Chang, Chaplot,
  Dosovitskiy, Gupta, Koltun, Kosecka, Malik, Mottaghi, Savva
  et~al.}]{Anderson:2018:Evaluation}
Peter Anderson, Angel Chang, Devendra~Singh Chaplot, Alexey Dosovitskiy,
  Saurabh Gupta, Vladlen Koltun, Jana Kosecka, Jitendra Malik, Roozbeh
  Mottaghi, Manolis Savva, et~al. 2018{\natexlab{a}}.
\newblock On evaluation of embodied navigation agents.
\newblock {\em arXiv preprint arXiv:1807.06757\/} .

\bibitem[{Anderson et~al.(2018{\natexlab{b}})Anderson, Wu, Teney, Bruce,
  Johnson, S{\"u}nderhauf, Reid, Gould, and van~den Hengel}]{Anderson:2018:VLN}
Peter Anderson, Qi~Wu, Damien Teney, Jake Bruce, Mark Johnson, Niko
  S{\"u}nderhauf, Ian Reid, Stephen Gould, and Anton van~den Hengel.
  2018{\natexlab{b}}.
\newblock Vision-and-language navigation: Interpreting visually-grounded
  navigation instructions in real environments.
\newblock In {\em Proceedings of the IEEE Conference on Computer Vision and
  Pattern Recognition (CVPR)\/}.

\bibitem[{{Antol} et~al.(2015){Antol}, {Agrawal}, {Lu}, {Mitchell}, {Batra},
  {Zitnick}, and {Parikh}}]{7410636}
S.~{Antol}, A.~{Agrawal}, J.~{Lu}, M.~{Mitchell}, D.~{Batra}, C.~L. {Zitnick},
  and D.~{Parikh}. 2015.
\newblock {VQA}: Visual question answering.
\newblock In {\em 2015 IEEE International Conference on Computer Vision
  (ICCV)\/}. pages 2425--2433.

\bibitem[{Bain and Sammut(1999)}]{Bain:1999:Cloning}
Michael Bain and Claude Sammut. 1999.
\newblock A framework for behavioural cloning.
\newblock In {\em Machine Intelligence 15, Intelligent Agents [St. Catherine's
  College, Oxford, July 1995]\/}. Oxford University, Oxford, UK, UK, pages
  103--129.

\bibitem[{Baldridge et~al.(2018)Baldridge, Bedrax-Weiss, Luong, Narayanan,
  Pang, Pereira, Soricut, Tseng, and Zhang}]{baldridge-etal-2018-points}
Jason Baldridge, Tania Bedrax-Weiss, Daphne Luong, Srini Narayanan, Bo~Pang,
  Fernando Pereira, Radu Soricut, Michael Tseng, and Yuan Zhang. 2018.
\newblock Points, paths, and playscapes: Large-scale spatial language
  understanding tasks set in the real world.
\newblock In {\em Proceedings of the First International Workshop on Spatial
  Language Understanding\/}. Association for Computational Linguistics, New
  Orleans, pages 46--52.

\bibitem[{Bisk et~al.(2018)Bisk, Shih, Choi, and Marcu}]{bisk:etal:2018}
Yonatan Bisk, Kevin Shih, Yejin Choi, and Daniel Marcu. 2018.
\newblock Learning interpretable spatial operations in a rich 3d blocks world.
\newblock In {\em Proceedings of the Thirty-Second Conference on Artificial
  Intelligence (AAAI-18)\/}. New Orleans, USA.

\bibitem[{Blukis et~al.(2018)Blukis, Misra, Knepper, and
  Artzi}]{Blukis:18visit-predict}
Valts Blukis, Dipendra Misra, Ross~A. Knepper, and Yoav Artzi. 2018.
\newblock Mapping navigation instructions to continuous control actions with
  position visitation prediction.
\newblock In {\em Proceedings of the Conference on Robot Learning\/}.

\bibitem[{Chen et~al.(2019)Chen, Suhr, Misra, and Artzi}]{Chen19:touchdown}
Howard Chen, Alane Suhr, Dipendra Misra, and Yoav Artzi. 2019.
\newblock Touchdown: Natural language navigation and spatial reasoning in
  visual street environments.
\newblock In {\em Conference on Computer Vision and Pattern Recognition\/}.

\bibitem[{Cirik et~al.(2018)Cirik, Zhang, and Baldridge}]{cirik2018following}
Volkan Cirik, Yuan Zhang, and Jason Baldridge. 2018.
\newblock Following formulaic map instructions in a street simulation
  environment.
\newblock {\em NIPS Visually Grounded Interaction and Language Workshop\/} .

\bibitem[{Daftry et~al.(2016)Daftry, Bagnell, and
  Hebert}]{Daftry:2016:TransferablePolicies}
Shreyansh Daftry, J.~Andrew Bagnell, and Martial Hebert. 2016.
\newblock Learning transferable policies for monocular reactive {MAV} control.
\newblock In {\em International Symposium on Experimental Robotics\/}.
  Springer, pages 3--11.

\bibitem[{Das et~al.(2017)Das, Kottur, Gupta, Singh, Yadav, Moura, Parikh, and
  Batra}]{visdial}
Abhishek Das, Satwik Kottur, Khushi Gupta, Avi Singh, Deshraj Yadav,
  Jos\'e~M.F. Moura, Devi Parikh, and Dhruv Batra. 2017.
\newblock {V}isual {D}ialog.
\newblock In {\em Proceedings of the IEEE Conference on Computer Vision and
  Pattern Recognition (CVPR)\/}.

\bibitem[{de~Vries et~al.(2018)de~Vries, Shuster, Batra, Parikh, Weston, and
  Kiela}]{DBLP:journals/corr/abs-1807-03367}
Harm de~Vries, Kurt Shuster, Dhruv Batra, Devi Parikh, Jason Weston, and Douwe
  Kiela. 2018.
\newblock Talk the walk: Navigating new york city through grounded dialogue.
\newblock {\em CoRR\/} abs/1807.03367.

\bibitem[{Dijkstra(1959)}]{dijkstra1959note}
Edsger~W Dijkstra. 1959.
\newblock A note on two problems in connexion with graphs.
\newblock {\em Numerische mathematik\/} 1(1):269--271.

\bibitem[{{Donahue} et~al.(2017){Donahue}, {Hendricks}, {Rohrbach},
  {Venugopalan}, {Guadarrama}, {Saenko}, and {Darrell}}]{7558228}
J.~{Donahue}, L.~A. {Hendricks}, M.~{Rohrbach}, S.~{Venugopalan},
  S.~{Guadarrama}, K.~{Saenko}, and T.~{Darrell}. 2017.
\newblock Long-term recurrent convolutional networks for visual recognition and
  description.
\newblock {\em IEEE Transactions on Pattern Analysis and Machine
  Intelligence\/} 39(4):677--691.

\bibitem[{{Fang} et~al.(2015){Fang}, {Gupta}, {Iandola}, {Srivastava}, {Deng},
  {Dollár}, {Gao}, {He}, {Mitchell}, {Platt}, {Zitnick}, and
  {Zweig}}]{7298754}
H.~{Fang}, S.~{Gupta}, F.~{Iandola}, R.~K. {Srivastava}, L.~{Deng},
  P.~{Dollár}, J.~{Gao}, X.~{He}, M.~{Mitchell}, J.~C. {Platt}, C.~L.
  {Zitnick}, and G.~{Zweig}. 2015.
\newblock From captions to visual concepts and back.
\newblock In {\em 2015 IEEE Conference on Computer Vision and Pattern
  Recognition (CVPR)\/}. pages 1473--1482.

\bibitem[{Fried et~al.(2018)Fried, Hu, Cirik, Rohrbach, Andreas, Morency,
  Berg-Kirkpatrick, Saenko, Klein, and Darrell}]{Fried:2018:Speaker}
Daniel Fried, Ronghang Hu, Volkan Cirik, Anna Rohrbach, Jacob Andreas,
  Louis-Philippe Morency, Taylor Berg-Kirkpatrick, Kate Saenko, Dan Klein, and
  Trevor Darrell. 2018.
\newblock Speaker-{F}ollower models for {V}ision-and-{L}anguage {N}avigation.
\newblock In {\em Neural Information Processing Systems (NeurIPS)\/}.

\bibitem[{Hochreiter and Schmidhuber(1997)}]{Hochreiter:1997:LSTM}
Sepp Hochreiter and J\"{u}rgen Schmidhuber. 1997.
\newblock Long short-term memory.
\newblock {\em Neural Comput.\/} 9(8):1735--1780.

\bibitem[{Huang et~al.(2019)Huang, Jain, Mehta, Baldridge, and
  Ie}]{huang-etal-splu-robonlp-2019}
Haoshuo Huang, Vihan Jain, Harsh Mehta, Jason Baldridge, and Eugene Ie. 2019.
\newblock Multi-modal discriminative model for vision-and-language navigation.
\newblock In {\em Proceedings of the Combined Workshop on Spatial Language
  Understanding and Grounded Communication for Robotics (SpLU-RoboNLP-2019)\/}.
  Association for Computational Linguistics, Minneapolis.

\bibitem[{Kruijff et~al.(2014)Kruijff, Kruijff-Korbayova, Keshavdas,
  Larochelle, Janicek, Colas, Liu, Pomerleau, Siegwart, Neerincx, Looije,
  Smets, Mioch, Diggelen, Pirri, Gianni, Ferri, Menna, Worst, and
  Hlavac}]{tradr-project}
G.J.M. Kruijff, Ivana Kruijff-Korbayova, Shanker Keshavdas, Benoit Larochelle,
  Miroslav Janicek, Francis Colas, Ming Liu, FranÃ§ois Pomerleau, Roland
  Siegwart, Mark Neerincx, Rosemarijn Looije, Nanja Smets, Tina Mioch, Jurriaan
  Diggelen, Fiora Pirri, Mario Gianni, Federico Ferri, Matteo Menna, Rainer
  Worst, and Vaclav Hlavac. 2014.
\newblock Designing, developing, and deploying systems to support human-robot
  teams in disaster response.
\newblock {\em Advanced Robotics\/} 28.

\bibitem[{{Kruijff-Korbayová} et~al.(2016){Kruijff-Korbayová}, {Freda},
  {Gianni}, {Ntouskos}, {Hlaváč}, {Kubelka}, {Zimmermann}, {Surmann},
  {Dulic}, {Rottner}, and {Gissi}}]{tradr-italy}
I.~{Kruijff-Korbayová}, L.~{Freda}, M.~{Gianni}, V.~{Ntouskos}, V.~{Hlaváč},
  V.~{Kubelka}, E.~{Zimmermann}, H.~{Surmann}, K.~{Dulic}, W.~{Rottner}, and
  E.~{Gissi}. 2016.
\newblock Deployment of ground and aerial robots in earthquake-struck amatrice
  in italy (brief report).
\newblock In {\em 2016 IEEE International Symposium on Safety, Security, and
  Rescue Robotics (SSRR)\/}. pages 278--279.

\bibitem[{Ma et~al.(2019)Ma, Lu, Wu, Alregib, Kira, Socher, and
  Xiong}]{Ma:2019:SelfMonitoringAgent}
Chih-Yao Ma, Jiasen Lu, Zuxuan Wu, Ghassan Alregib, Zsolt Kira, Richard Socher,
  and Caiming Xiong. 2019.
\newblock Self-monitoring navigation agent via auxiliary progress estimation.
\newblock In {\em Proceedings of the International Conference on Learning
  Representations (ICLR)\/}.

\bibitem[{Macmahon et~al.(2006)Macmahon, Stankiewicz, and
  Kuipers}]{Macmahon06walkthe}
Matt Macmahon, Brian Stankiewicz, and Benjamin Kuipers. 2006.
\newblock Walk the talk: Connecting language, knowledge, action in route
  instructions.
\newblock In {\em Proceedings of the National Conference on Artificial
  Intelligence (AAAI)\/}. pages 1475--1482.

\bibitem[{Misra et~al.(2018)Misra, Bennett, Blukis, Niklasson, Shatkhin, and
  Artzi}]{misra-etal-2018-mapping}
Dipendra Misra, Andrew Bennett, Valts Blukis, Eyvind Niklasson, Max Shatkhin,
  and Yoav Artzi. 2018.
\newblock Mapping instructions to actions in 3{D} environments with visual goal
  prediction.
\newblock In {\em Proceedings of the 2018 Conference on Empirical Methods in
  Natural Language Processing\/}. Association for Computational Linguistics,
  Brussels, Belgium, pages 2667--2678.

\bibitem[{Pennington et~al.(2014)Pennington, Socher, and
  Manning}]{Pennington:2014:GloVe}
Jeffrey Pennington, Richard Socher, and Christopher Manning. 2014.
\newblock Glove: Global vectors for word representation.
\newblock In {\em Proceedings of the 2014 Conference on Empirical Methods in
  Natural Language Processing (EMNLP)\/}. Association for Computational
  Linguistics, pages 1532--1543.

\bibitem[{Schuster and Paliwal(1997)}]{Schuster:1997:BiRNN}
M.~Schuster and K.K. Paliwal. 1997.
\newblock Bidirectional recurrent neural networks.
\newblock {\em Trans. Sig. Proc.\/} 45(11):2673--2681.

\bibitem[{Shah et~al.(2018)Shah, Fiser, Faust, Kew, and
  Hakkani-Tur}]{follownet}
Pararth Shah, Marek Fiser, Aleksandra Faust, Chase Kew, and Dilek Hakkani-Tur.
  2018.
\newblock Follow{N}et: Robot navigation by following natural language
  directions with deep reinforcement learning.
\newblock In {\em Third Machine Learning in Planning and Control of Robot
  Motion Workshop at ICRA\/}.

\bibitem[{{Sko{\v c}aj} et~al.(2016){Sko{\v c}aj}, {Vre{\v c}ko}, {Mahni{\v
  c}}, {Jan{\'{\i}}{\v c}ek}, {Kruijff}, {Hanheide}, {Hawes}, {Wyatt},
  {Keller}, {Zhou}, {Zillich}, and {Kristan}}]{skovaj:etal:2016}
D.~{Sko{\v c}aj}, A.~{Vre{\v c}ko}, M.~{Mahni{\v c}}, M.~{Jan{\'{\i}}{\v c}ek},
  G.-J.~M. {Kruijff}, M.~{Hanheide}, N.~{Hawes}, J.~L. {Wyatt}, T.~{Keller},
  K.~{Zhou}, M.~{Zillich}, and M.~{Kristan}. 2016.
\newblock {An integrated system for interactive continuous learning of
  categorical knowledge}.
\newblock {\em Journal of Experimental \& Theoretical Artificial
  Intelligence\/} 28:823--848.

\bibitem[{Thomason et~al.(2019)Thomason, Gordon, and Bisk}]{thomason:naacl19}
Jesse Thomason, Daniel Gordon, and Yonatan Bisk. 2019.
\newblock Shifting the baseline: Single modality performance on visual
  navigation \& {QA}.
\newblock In {\em Conference of the North American Chapter of the Association
  for Computational Linguistics (NAACL)\/}.

\bibitem[{Thomason et~al.(2018)Thomason, Sinapov, Mooney, and
  Stone}]{thomason:aaai18}
Jesse Thomason, Jivko Sinapov, Raymond Mooney, and Peter Stone. 2018.
\newblock Guiding exploratory behaviors for multi-modal grounding of linguistic
  descriptions.
\newblock In {\em Proceedings of the Thirty-Second AAAI Conference on
  Artificial Intelligence (AAAI-18)\/}.

\bibitem[{Vinyals et~al.(2015)Vinyals, Toshev, Bengio, and
  Erhan}]{Vinyals2015ShowAT}
Oriol Vinyals, Alexander Toshev, Samy Bengio, and Dumitru Erhan. 2015.
\newblock Show and tell: A neural image caption generator.
\newblock {\em 2015 IEEE Conference on Computer Vision and Pattern Recognition
  (CVPR)\/} pages 3156--3164.

\bibitem[{Wang et~al.(2018)Wang, Chen, Wu, Wang, and Wang}]{Wang2018VideoCV}
Xin Wang, Wenhu Chen, Jiawei Wu, Yuan-Fang Wang, and William~Yang Wang. 2018.
\newblock Video captioning via hierarchical reinforcement learning.
\newblock {\em 2018 IEEE/CVF Conference on Computer Vision and Pattern
  Recognition\/} pages 4213--4222.

\bibitem[{Wang et~al.(2019)Wang, Huang, {\c{C}}elikyilmaz, Gao, Shen, Wang,
  Wang, and Zhang}]{Wang:2018:RCM}
Xin Wang, Qiuyuan Huang, Asli {\c{C}}elikyilmaz, Jianfeng Gao, Dinghan Shen,
  Yuan{-}Fang Wang, William~Yang Wang, and Lei Zhang. 2019.
\newblock Reinforced cross-modal matching and self-supervised imitation
  learning for vision-language navigation.
\newblock In {\em Proceedings of the IEEE Conference on Computer Vision and
  Pattern Recognition (CVPR)\/}.

\bibitem[{Williams et~al.(2018)Williams, Gopalan, Rhee, and
  Tellex}]{WilliamsGRT18}
Edward~C. Williams, Nakul Gopalan, Mina Rhee, and Stefanie Tellex. 2018.
\newblock Learning to parse natural language to grounded reward functions with
  weak supervision.
\newblock In {\em 2018 {IEEE} International Conference on Robotics and
  Automation, {ICRA} 2018, Brisbane, Australia, May 21-25, 2018\/}. pages 1--7.

\bibitem[{Williams(1992)}]{Williams:1992:PolicyGradient}
Ronald~J. Williams. 1992.
\newblock Simple statistical gradient-following algorithms for connectionist
  reinforcement learning.
\newblock {\em Machine Learning\/} 8(3):229--256.

\bibitem[{Yan et~al.(2018)Yan, Misra, Bennett, Walsman, Bisk, and
  Artzi}]{Yan:18}
Claudia Yan, Dipendra Misra, Andrew Bennett, Aaron Walsman, Yonatan Bisk, and
  Yoav Artzi. 2018.
\newblock {CHALET: Cornell House Agent Learning Environment}.
\newblock {\em CoRR\/} abs/1801.07357.

\bibitem[{Yang et~al.(2016)Yang, He, Gao, Deng, and Smola}]{Yang2016StackedAN}
Zichao Yang, Xiaodong He, Jianfeng Gao, Li~Deng, and Alexander~J. Smola. 2016.
\newblock Stacked attention networks for image question answering.
\newblock {\em 2016 IEEE Conference on Computer Vision and Pattern Recognition
  (CVPR)\/} pages 21--29.

\bibitem[{Yu et~al.(2016)Yu, Wang, Huang, Yang, and Xu}]{YuCVPR2016}
Haonan Yu, Jiang Wang, Zhiheng Huang, Yi~Yang, and Wei Xu. 2016.
\newblock Video paragraph captioning using hierarchical recurrent neural
  networks.
\newblock In {\em 2016 {IEEE} Conference on Computer Vision and Pattern
  Recognition, {CVPR} 2016, Las Vegas, NV, USA, June 27-30, 2016\/}. pages
  4584--4593.

\end{thebibliography}
% \bibliographystyle{acl_natbib}

\end{document}